\documentclass[twocolumn]{article}

\usepackage{algorithm}
\usepackage{algpseudocode}
\usepackage[utf8]{inputenc}
\usepackage{amsmath}
\usepackage{graphicx}
\usepackage{hyperref}
\usepackage{fancybox}
\usepackage{tabularx}
\usepackage{listings}
\usepackage{xcolor}
\usepackage{placeins}
\usepackage{amssymb}
\usepackage[skip=2pt,font=scriptsize]{caption}
\usepackage[backend=biber,style=numeric]{biblatex}
\begin{document}
\title{\textbf{BERT-LSH: Reducing Absolute Compute For Attention}}
\author{Zezheng Li and Kingston Yip\\
\texttt{New York University}
}
\date{}

\maketitle

\begin{abstract}
This study introduces a novel \texttt{BERT-LSH} model that incorporates Locality Sensitive Hashing (LSH) to approximate the attention mechanism in the \texttt{BERT} architecture. We examine the computational efficiency and performance of this model compared to a standard baseline \texttt{BERT} model. Our findings reveal that \texttt{BERT-LSH} significantly reduces computational demand for the self attention layer while unexpectedly outperforming the baseline model in pretraining and fine-tuning tasks. These results suggest that the LSH-based attention mechanism not only offers computational advantages but also may enhance the model's ability to generalize from its training data. \Ovalbox{\href{https://github.com/leo4life2/algoml-final}{https://github.com/leo4life2/algoml-final}}

\end{abstract}

\section{Introduction}

Transformer-based models such as \texttt{BERT} \cite{devlin2019bert} have markedly advanced NLP. However, their quadratic complexity in the self-attention mechanism leads to computational inefficiencies, especially with longer sequences \cite{vaswani2023attention}.

To address this, the Reformer \cite{kitaev2020reformer} modifies the attention mechanism by utilizing locality-sensitive hashing (LSH) \cite{andoni2006near} to reduce complexity. It achieves this by merging the query (\textbf{Q}) and key (\textbf{K}) matrices into a single \textbf{QK} matrix and applying LSH to bucket similar items together, thus limiting the attention computation to items within the same buckets. This approach significantly reduces memory usage and computational load.

Our \texttt{BERT-LSH} model diverges from the Reformer by retaining the distinct \textbf{Q} and \textbf{K} matrices of the original \texttt{BERT} architecture, preserving the nuanced distinction between queries and keys in attention computation. \texttt{BERT-LSH} applies LSH to both \textbf{Q} and \textbf{K} matrices independently, ensuring that only the rows that hash into common buckets across all hash functions are considered for subsequent dot product calculations. This method not only retains the expressiveness of BERT's attention mechanism but also leverages the computational efficiency of LSH, striking a balance between model performance and resource utilization.

By focusing on the intersection of LSH buckets from \textbf{Q} and \textbf{K} matrices, \texttt{BERT-LSH} introduces a novel way to approximate full attention that is both computationally efficient and sensitive to the distinct semantics of queries and keys. This approach is particularly advantageous for applications requiring the deployment of powerful NLP models in resource-constrained environments. Our study evaluates the effectiveness of \texttt{BERT-LSH} in maintaining performance while offering computational benefits, thus contributing to the development of more scalable and practical NLP solutions.

\section{Methodology}

\subsection{Locality Sensitive Hashing Implementation} \label{methodology:locality}

Our SimHash-based Locality Sensitive Hashing (LSH) scheme is an optimized vectorized adaptation of a traditional LSH scheme designed for cosine similarity. The scheme is structured around $r$ bands and leverages a set of random vectors $\mathbf{g}_1, \ldots, \mathbf{g}_r \in \mathbb{R}^d$, each with entries drawn from a standard normal distribution $\mathcal{N}(0,1)$. A uniform random hash function, denoted as $f:\{-1,1\}^r \rightarrow\{1, \ldots, m\}$, maps binary vectors to scalar hash values. The hash function $h: \mathbb{R}^d \rightarrow\{1, \ldots, m\}$ for a vector $\mathbf{x}$ is defined as:
\begin{equation*}
h(\mathbf{x})=f\left(\left[\operatorname{sign}\left(\left\langle\mathbf{g}_1, \mathbf{x}\right\rangle\right), \ldots, \operatorname{sign}\left(\left\langle\mathbf{g}_r, \mathbf{x}\right\rangle\right)\right]\right)
\end{equation*}
Our implementation expands upon this by incorporating multiple hash functions, thus increasing the likelihood of collision for similar vectors across different hash functions:
\begin{equation*}
h_i(\mathbf{x})=f_i\left(\left[\operatorname{sign}\left(\left\langle\mathbf{g}_{i1}, \mathbf{x}\right\rangle\right), \ldots, \operatorname{sign}\left(\left\langle\mathbf{g}_{ir}, \mathbf{x}\right\rangle\right)\right]\right)
\end{equation*}

The uniform hash function converts a signed binary vector to a scalar by summing the products of random coefficients $C_i \in \{1, \ldots, m\}$ and the positive entries of the vector to produce a scalar hash value:
\begin{equation*}
f(\mathbf{s}) = \left(\sum_{i=1}^{r} \mathbf{1}_{\{s_i > 0\}} \cdot s_i \cdot C_i \right) \mod m
\end{equation*}
where $\mathbf{s}$ is the signed binary vector and $s_i$ is the $i$-th element of $\mathbf{s}$.

In practice, the hash function is applied to a matrix of vectors to produce a batch of hash values. This process involves first computing the sign hash of each vector, followed by the application of the uniform hash function across all bands and vectors. This vectorized operation allows us to efficiently generate a hash table that maps vectors to buckets.

To populate the bucket table, we employ a vectorized approach to compute the hash values for all vectors using all hash functions. Each vector $\mathbf{v}$ is hashed, and the resulting hash values are used to update the bucket table, with vectors that share the same hash value being grouped together.

A collision matrix $\mathbf{M}$ is constructed to record vector collisions, suggesting potential similarity. For vector set $\mathbf{V}$, the collision matrix is:
\begin{equation*}
M_{ij} = 
\begin{cases}
1, & \text{if } \exists h_k \in \mathcal{H} : h_k(\mathbf{v}_i) = h_k(\mathbf{v}_j) \\
0, & \text{otherwise}
\end{cases}
\end{equation*}
Here, $\mathbf{v}_i$ and $\mathbf{v}_j$ represent the $i$-th and $j$-th vectors in $\mathbf{V}$, respectively, and $\mathcal{H}$ denotes the set of all hash functions employed to generate the hash codes for the vectors. 

In the context of our \texttt{BERT-LSH} model, we LSH to approximate the attention mechanism, which traditionally relies on the computation of pairwise similarity between query ($\mathbf{Q}$) and key ($\mathbf{K}$) vectors. To facilitate this approximation, $\mathbf{Q}$ and $\mathbf{K}$ are stacked vertically to form a combined matrix $\mathbf{M}$, where the top half consists of the query vectors and the bottom half comprises the key vectors. This configuration allows us to apply LSH to the entire matrix $\mathbf{M}$ at once, thereby capturing the collisions between query and key vectors within a single computational framework. The collisions, which indicate potential relevance between queries and keys, are represented in a collision matrix. Within this matrix, the subset that is of particular interest to us is the top right quadrant, as it contains the collision information between the vectors of $\mathbf{Q}$ and $\mathbf{K}$. This relevant submatrix of $\mathbf{M}$, denoted as $\mathbf{C}$, is shown below:

\[
\mathbf{M} = \left[ \begin{array}{c|c}
\mathbf{I_{QQ}} & \mathbf{C} \\
\hline
\mathbf{C}^\top & \mathbf{I_{KK}}
\end{array} \right]
\]

$\mathbf{C}$ represents the collisions between query and key vectors and is of interest to us. $\mathbf{I}_{QQ}$ and $\mathbf{I}_{KK}$ are matrices that represent collisions within the query and key vectors, which is irrelevant for our use case. Our attention to $\mathbf{C}$ is driven by the hypothesis that the LSH-induced collisions between $\mathbf{Q}$ and $\mathbf{K}$ can serve as a proxy for the computationally intensive pairwise similarity measures, thus providing a basis for the efficient computation of the attention scores within the \texttt{BERT-LSH} framework.

In our SimHash-based Locality Sensitive Hashing (LSH) implementation, we can quantify the probability of collision between vectors, which is a pivotal factor in the effectiveness of the hashing scheme. The probability of collision depends on the angle $\theta$ between two vectors in the high-dimensional space.

Given a single hyperplane defined by a random vector $\mathbf{g}$, the probability that two vectors $\mathbf{a}$ and $\mathbf{b}$ fall on the same side of the hyperplane (i.e., resulting in a collision) is $1 - \frac{\theta}{\pi}$, where $\theta$ is the angle between $\mathbf{a}$ and $\mathbf{b}$. Considering $r$ bands, each with its own random vector, the probability that $\mathbf{a}$ and $\mathbf{b}$ lie on the same side of the hyperplane is:

\begin{equation*}
P(\langle\mathbf{g}, \mathbf{a}\rangle=\langle\mathbf{g}, \mathbf{b}\rangle) = \left(1 - \frac{\theta}{\pi}\right)^r
\end{equation*}

However, even if the vectors are not similar, they may still be hashed to the same bucket due to the random uniform hash function $f$. This introduces an additional probability of $\frac{1}{m}$, where $m$ is the number of buckets.

\begin{equation*}
P(h(\mathbf{a}) \neq h(\mathbf{b})) = 1 - \left[ \left(1 - \frac{\theta}{\pi}\right)^r + \frac{1}{m} \right]
\end{equation*}

With $n$ independent hash functions, the probability that vectors $\mathbf{a}$ and $\mathbf{b}$ do not collide in any bucket across all hash functions is the product of the individual probabilities of no collision for each hash function. Hence, the probability of at least one collision over all $n$ hash functions is:

\begin{align*}
P(\exists h_k \in \mathcal{H} : h_k(\mathbf{a}) = h_k(\mathbf{b})) = \\
1 - \left(1 - \left[\left(1 - \frac{\theta}{\pi}\right)^r + \frac{1}{m}\right]\right)^n
\end{align*}

As we increase the number of hash functions \( n \), the likelihood of capturing genuine similarities (true positives) rises, albeit at the risk of also increasing false positives. Conversely, increasing the number of bands \( r \) tends to reduce false positives but may lead to a higher rate of false negatives, thereby necessitating a careful balance to optimize the trade-off between the sensitivity and specificity of our LSH scheme in the \texttt{BERT-LSH} model.

\subsection{Model Implementation}

In our study, we implemented and trained two distinct models based on the \texttt{BERT} architecture, utilizing the Transformers library provided by Huggingface for ease of use and reproducibility.

\subsubsection{BERT Baseline Model}

The first model we implemented served as the control in our experiments and is a standard representation of the original \texttt{BERT} model, which is a standard transformer encoder tower. We utilized the \texttt{BertModel} class directly from the Huggingface's Transformers library to instantiate this model. The configuration of this model was chosen to resemble the \texttt{BERT-Tiny} architecture, which is a scaled-down version of the original BERT, designed to be lightweight and trainable on personal computing resources as it only has around 4.7 million parameters. The specific parameters for our baseline \texttt{BERT} model were as follows: a hidden layer size of 128, a total of 2 hidden layers, 2 attention heads, and an intermediate size of 512 for the feed-forward layers. This configuration was selected to ensure that the model could be effectively trained on a personal computer equipped with a single RTX 4090 GPU.

\subsubsection{BERT-LSH Model}

The second model is a novel \texttt{BERT-LSH} model, which incorporates a key modification to the self-attention mechanism of the original \texttt{BERT} model. The alteration was minimal yet impactful: we substituted the batched matrix multiplication for the $QK^T$ operation with our LSH-based attention mechanism. To achieve this, we first initialized an empty attention score matrix. We then iterated over all batches and heads, applying the LSH algorithm to the $Q$ and $K$ matrices within the same batch and head to identify collisions. When a collision was detected—signifying that a pair of vectors from $Q$ and $K$ landed in the same hash bucket in at least one hash function—we calculated the dot product to determine their attention score. These scores were then symmetrically populated into the attention score matrix at the corresponding indices.

\begin{algorithm}[ht]
\caption{LSH-based attention mechanism algorithm}
\begin{algorithmic}[1] 
\State Initialize an empty attention score matrix $A$.
\For{each batch $b$}
    \For{each head $h$}
        \State $\mathbf{C} \gets \text{LSH}(Q_{b,h}, K_{b,h})$ \Comment{Get the collision matrix for the current head}
        \ForAll{$(i, j) \in \{ (x, y) \mid \mathbf{C}[x, y] = 1 \}$}
            \State $A_{b,h}[i,j] \gets \langle q_i, k_j \rangle$
            \State $A_{b,h}[j,i] \gets A_{b,h}[i,j]$
        \EndFor
    \EndFor
\EndFor
\end{algorithmic}
\end{algorithm}

This design choice resulted in an attention score matrix that is symmetric along its diagonal, reflecting the bidirectional nature of BERT. Initially, there was concern that this symmetry might adversely affect the model's performance. However, considering BERT's inherent bidirectionality and the use of positional encodings to embed token order information, we deemed the approach valid. The configuration parameters for the \texttt{BERT-LSH} model mirrored those of the baseline model, ensuring a fair comparison in terms of model capacity and computational demand.

The \texttt{BERT-LSH} model represents a significant departure from traditional attention mechanisms by leveraging the efficiency of LSH to approximate attention scores. This innovation has the potential to reduce computational complexity significantly while maintaining the representational power of the \texttt{BERT} architecture.

\subsection{Measuring Computational Efficiency}

To evaluate the computational efficiency of our \texttt{BERT-LSH} model relative to the baseline \texttt{BERT} model, we conducted a series of measurements focusing on the computational demands and execution time of the attention mechanisms. Our analysis was based on an LSH configuration with 2 bands, a table size of 64, and a single hash function. The input to both models consisted of a single random input sequence with a sequence length of 10. The following metrics were assessed:

\begin{enumerate}
    \item \textbf{Kilo Floating Point Operations (KFLOPs)}: We quantified the computational complexity in terms of the number of floating point operations (FLOPs), specifically measuring in thousands (KFLOPs), for a single execution of the attention mechanism across all batches and heads. Utilizing PyTorch's built-in profiler, we captured the FLOPs for both the baseline BERT's $QK^T$ multiplication and our LSH approximation. This direct measurement provided an objective comparison of the raw computational workload between the two approaches.
    
    \item \textbf{Dot Product Calculations}: A fundamental operation in the attention mechanism is the dot product, which we measured to compare the full self-attention with our LSH approximation. For the baseline BERT, the number of dot products corresponds to the product of batch size, number of heads, and the square of the sequence length, resulting in $1 \times 2 \times 10 \times 10 = 200$ dot products. In contrast, the LSH-based attention only computes dot products for pairs of query and key vectors that have collided post-LSH processing. Due to the stochastic nature of LSH, we performed 100 executions and calculated the average number of dot products to obtain a representative figure for the LSH attention.
    
    \item \textbf{Average Execution Time}: The efficiency of an operation is also gauged by its execution time. We executed the attention mechanism of both the baseline \texttt{BERT} and \texttt{BERT-LSH} models 1000 times to determine the average execution time for the $QK^T$ operation and its LSH approximation, respectively. This provided us with a direct comparison of the time efficiency between the conventional attention mechanism and our proposed LSH-based alternative.
\end{enumerate}

These measurements provide insight into the trade-offs between computational complexity and efficiency, crucial for assessing \texttt{BERT-LSH}'s practicality in computationally constrained real-world applications.

\subsection{Pretraining}

Pretraining is a critical step in the development of language models like BERT, as it allows the model to learn the intricacies and patterns of the language from a large corpus of text. For our \texttt{BERT} models, we adhered to the pretraining procedure established by the original BERT, which utilizes the Masked Language Modeling (MLM) technique.

\subsubsection{Masked Language Modeling (MLM)}

MLM is a pretraining objective that involves randomly masking out tokens in the input sentences and training the model to predict the original identity of the masked tokens. This approach encourages the model to develop a deep contextual understanding of the language, as it has to rely on the surrounding words to infer the masked tokens. By leveraging the Huggingface Transformers library, we were able to transform our base \texttt{BERT} models into instances of the \texttt{BertForMaskedLM} class. This adaptation appends a prediction head on top of the base \texttt{BERT} architecture, equipping the model with the ability to predict the masked tokens during pretraining.

\subsubsection{Dataset and Training Procedure}

For pretraining our models, we selected the \texttt{BookCorpus} dataset \cite{Zhu_2015_ICCV}, a choice inspired by the dataset used by the original \texttt{BERT} authors. While they utilized both the \texttt{BookCorpus} and \texttt{Wikipedia} datasets, our computational resources dictated a more focused approach. We opted to use only the \texttt{BookCorpus} dataset, and due to resource constraints, we further narrowed our training data to the top 10\% of the rows, amounting to 7.4 million rows of text.

Our pretraining regimen spanned 3 epochs over the selected subset of the \texttt{BookCorpus} dataset. We allocated 1\% of the data to serve as a test set, which would later be used to evaluate the model's performance in terms of loss on unseen data. The training loss data was collected throughout the epochs.

\subsubsection{Training Duration and Evaluation}

Pretraining on a PC with an RTX 4090 GPU took roughly 1 hour and 20 minutes for both models. Post-training, we evaluated performance by computing the test set loss.

\subsection{GLUE SST-2 Fine-tuning}

The pretraining phase of our \texttt{BERT} models was followed by fine-tuning on specific tasks in the General Language Understanding Evaluation (GLUE) benchmark, notably the Stanford Sentiment Treebank (SST-2) task, to adapt the models to particular datasets and problems. The GLUE benchmark, a collection of various natural language understanding tasks, includes the SST-2 task, which focuses on sentiment analysis of movie reviews, requiring classification of each review's sentiment as positive or negative. This approach allowed us to evaluate our models' performance across classification tasks, including sentiment analysis, textual entailment, and similarity assessment.

\subsubsection{Fine-tuning Procedure}

To adapt our \texttt{BERT} models for the sequence classification task presented by SST-2, we utilized the \texttt{BertForSequenceClassification} class from the Huggingface Transformers library. This class is specifically designed for sequence classification tasks, adding a classification layer on top of the pretrained \texttt{BERT} model. We loaded our previously trained weights into this class, thereby preserving the language understanding developed during pretraining while adapting the model for sentiment classification.

The SST-2 dataset is available through Huggingface's \texttt{glue} dataset collection. However, the original authors of the SST-2 dataset have not publicly released the test split. Consequently, we trained our models on the entire provided training split over 3 epochs. To evaluate the models' performance post fine-tuning, we used the validation split as our test set.

\subsubsection{Training and Evaluation Metrics}

During the fine-tuning process, we recorded the training loss at each step to monitor the models' learning progress. Upon the completion of fine-tuning, the models were evaluated on the validation split, which served as our test set. We recorded the loss on this set to quantify the models' performance on data that they had not been exposed to during training. This loss, along with other key metrics such as accuracy, provided a comprehensive view of how well each model was able to generalize the sentiment analysis task to new data.

\subsection{SQuAD Fine-tuning}

After completing the pretraining and fine-tuning phases on the GLUE SST-2 dataset, we transitioned to the task of question answering, utilizing the Stanford Question Answering Dataset (SQuAD) to assess our models. Both \texttt{BERT-LSH} and Baseline \texttt{BERT} were fine-tuned on SQuAD, a challenging dataset comprising questions based on Wikipedia articles, with answers being text segments from the articles. As SQuAD lacks a test split on HuggingFace, we adopted the same strategy as with SST-2, using the provided validation split as our test set for evaluating the models' question-answering performance.

\subsubsection{Fine-tuning Procedure}

For the question answering task, we employed the \texttt{BertForQuestionAnswering} classes from the Huggingface library. These classes are specifically designed for the question answering task, integrating the necessary prediction heads on top of the \texttt{BERT} models to handle the extraction of answer spans from the text. We initialized these classes with the weights from our pretrained \texttt{BERT-LSH} and Baseline \texttt{BERT} models, respectively.

We accessed the \texttt{squad} dataset directly from the Huggingface datasets library. The fine-tuning process involved training the models on the `train' split of the dataset for a total of three epochs. This duration was chosen to balance between model performance and the risk of overfitting.

\subsubsection{Training and Evaluation Metrics}

Throughout the fine-tuning process, we again continuously tracked the training loss. Once fine-tuning was complete, we conducted an evaluation on the `validation' split. We recorded the final test loss to assess how well the models were performing on unseen data.

\section{Results \& Analysis}
Moving forward in our visualizations, \texttt{Bert-LSH }will be represented with \textcolor{red}{red} markers or lines, while \texttt{Baseline Bert} will be denoted in \textcolor{blue}{blue}, allowing for a clear distinction between the two in our comparisons. \textbf{Bolded} values indicate a better or faster performance when compared to the other model.

\subsection{Computational Efficiency Results}

\begin{figure}[t]
    \centering
    \includegraphics[width=\linewidth]{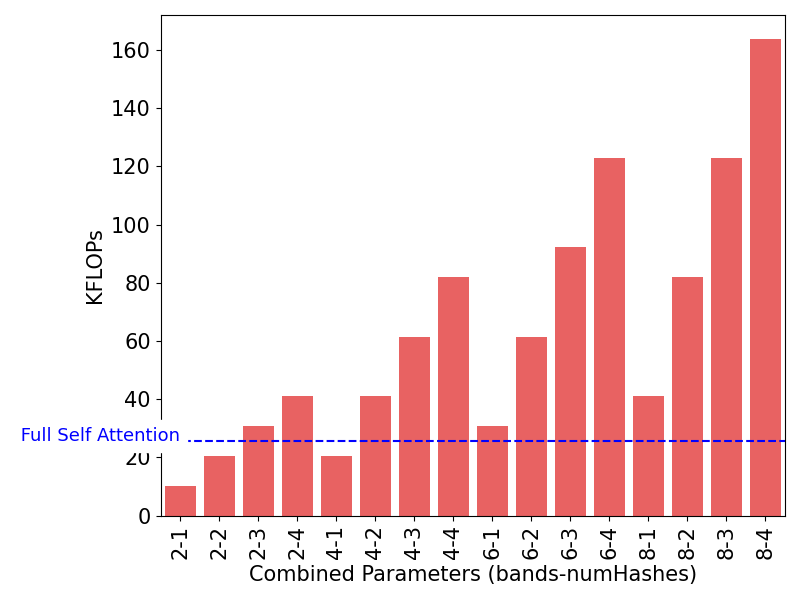}
    \caption{KFLOPs of the attention computation for different LSH configurations}
    \label{fig: kflops}
\end{figure}

In the LSH configuration of 2 bands, a table size of 64, and 1 hash function, \texttt{BERT-LSH} demonstrated a significant reduction in computational demand, using approximately 40\% of the KFLOPs required by full-self attention as shown in figure \ref{fig: kflops}. This efficiency was a key factor in opting for this specific LSH setup, as it has the potential of speeding up training and consuming less computational resources. Upon inspecting results of the PyTorch profiler, we discovered that a considerable amount of CPU time was dedicated to populating the zeros and initializing the collision matrix \textbf{M}\ref{methodology:locality}, but the overall reduction in computational requirements was substantial, enhancing the model's training efficiency for the self attention layer.

The computational efficiency of \texttt{BERT-LSH} was benchmarked against the baseline \texttt{BERT} model. The evaluation criteria included the KFLOPs necessary for a single forward pass, the average number of dot products during the attention calculation, and the average execution time for the matrix multiplication (\(QK^T\)) in the attention mechanism.

\begin{table}[h!]
\centering
\scriptsize 
\begin{tabularx}{\columnwidth}{Xcc} 
\hline
\textbf{Metric} & \textcolor{red}{\texttt{Baseline BERT}} & \textcolor{blue}{\texttt{BERT-LSH}} \\
\hline
KFLOPs/exec & 25.60 & \textbf{10.24} \\
Dot Prods (avg/100 execs) & 200 & \textbf{28.5} \\
Exec Time (avg/1k execs, s) & \textbf{1.22e-5} & 3.37e-4  \\
\hline
\end{tabularx}
\caption{Computational Efficiency Comparison between \texttt{BERT-LSH} and Baseline BERT}
\label{tab:efficiency}
\end{table}
In our examination of \texttt{BERT-LSH} across various LSH configurations, we found that computational complexity, measured in KFLOPs, varied linearly with the number of hash functions and bands but remained constant with changes in table size. The linear increase in KFLOPs with more hash functions results from the need to recalculate hashes for each vector with each additional function, directly impacting computational load. Table size, which dictates hash table capacity, doesn't affect the computation of hashes, and thus is left out of the plot. Moreover, increasing the number of bands leads to a proportional rise in KFLOPs due to additional dot product computations required for each dataset vector. These observations, supported by empirical data, are clearly illustrated in Figure \ref{fig: kflops}.

The observed slower execution time for BERT-LSH, despite reduced operations, was anticipated. Our LSH attention scheme successfully lowers time complexity, evident in fewer KFLOPs and dot products. However, the implementation lags behind the highly optimized PyTorch matrix operations. Our optimization for parallelism hits a bottleneck due to the linear nature of populating the bucket table and constructing the collision matrix, a process not as inherently parallelizable as standard matrix multiplication. With current GPUs excelling at parallel tasks, unless \texttt{BERT-LSH} can match this level of parallelism, it may not accelerate attention computations in practice.

\subsection{Pretraining Results}

Contrary to initial expectations, the \texttt{BERT-LSH} model not only achieved a lower training loss but also outperformed the baseline in evaluation metrics, indicating a better understanding of the training data. This outcome was surprising, as we hypothesized that the sparsity introduced by our LSH attention scheme might compromise the model's effectiveness. Nonetheless, the BERT-LSH's evaluation loss and accuracy improvements suggest that it has learned more effectively, possibly due to a more focused attention mechanism that filters out less relevant information, thus allowing for a more efficient learning process.

\begin{table}[ht!]
\centering
\scriptsize 
\begin{tabularx}{\columnwidth}{Xcc} 
\hline
\textbf{Metric} & \textcolor{red}{\texttt{Baseline BERT}} & \textcolor{blue}{\texttt{BERT-LSH}} \\
\hline
Epochs                        & 3.0                    & 3.0               \\
Training Loss                 & 5.6068                 & \textbf{5.5512}            \\
Evaluation Loss               & 5.4955                 & \textbf{5.3308}            \\
Evaluation Accuracy (\%)      & 15.05                  & \textbf{18.19}             \\
Perplexity                    & 243.59                 & \textbf{206.61}            \\
Training Runtime(s)          & \textbf{1285.25 }               & 4178.48           \\
Train Samples/s      & \textbf{549.27}                 & 168.95            \\
Train Steps/s        & \textbf{68.66}                  & 21.12             \\
Evaluation Runtime(s)        & \textbf{3.00}                   & 9.88              \\
Evaluation Samples/s & \textbf{874.29}                 & 265.79            \\
Evaluation Steps/s   & \textbf{109.58}                 & 33.31             \\
\hline
\end{tabularx}
\caption{Pretraining Benchmarks for \texttt{BERT} Baseline and \texttt{BERT-LSH} Models.}
\label{tab:training_benchmarks}
\end{table}
\FloatBarrier

Furthermore, the fact that the \texttt{BERT-LSH} model demonstrated superior performance on the test split indicates that our LSH approach did not lead to overfitting. Instead, it seems to have enhanced the model's ability to generalize from its training data.

\begin{figure}[htp]
    \centering
    \includegraphics[width=\linewidth]{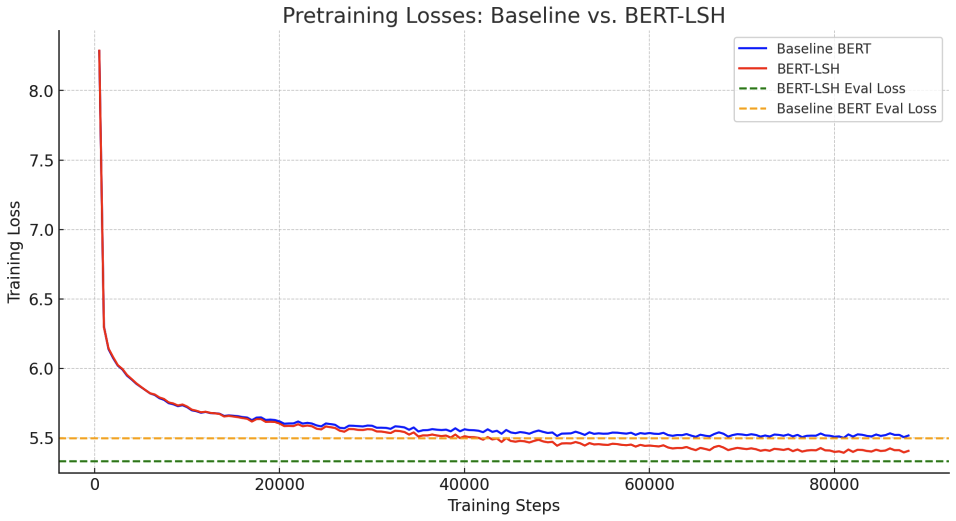}
    \caption{\textbf{Pretraining Losses}: \texttt{BERT-LSH} showed better Eval Loss than when compared to the Baseline \texttt{BERT} model during pretraining.}
    \label{fig:pretraining_loss}
\end{figure}

The lower loss on the test split, in particular, points to a better generalization capability when compared to the baseline \texttt{BERT} model. This suggests that the LSH-based attention mechanism may be capturing the underlying data distribution more effectively, enabling the model to perform better on unseen data. These results are promising and warrant further investigation into the potential benefits of LSH in large-scale language models.
 
However, these performance gains come at a significant cost in terms of total training times. The \texttt{BERT-LSH} model requires approximately three times longer to train than the baseline \texttt{BERT} model. Given that \texttt{BERT-LSH} was not optimized for parallel computation, this aligns with our expectations as PyTorch's matrix multiplications are highly parallelized making them much faster than our custom implementation.

\begin{figure}[htp!]
    \includegraphics[width=\columnwidth]{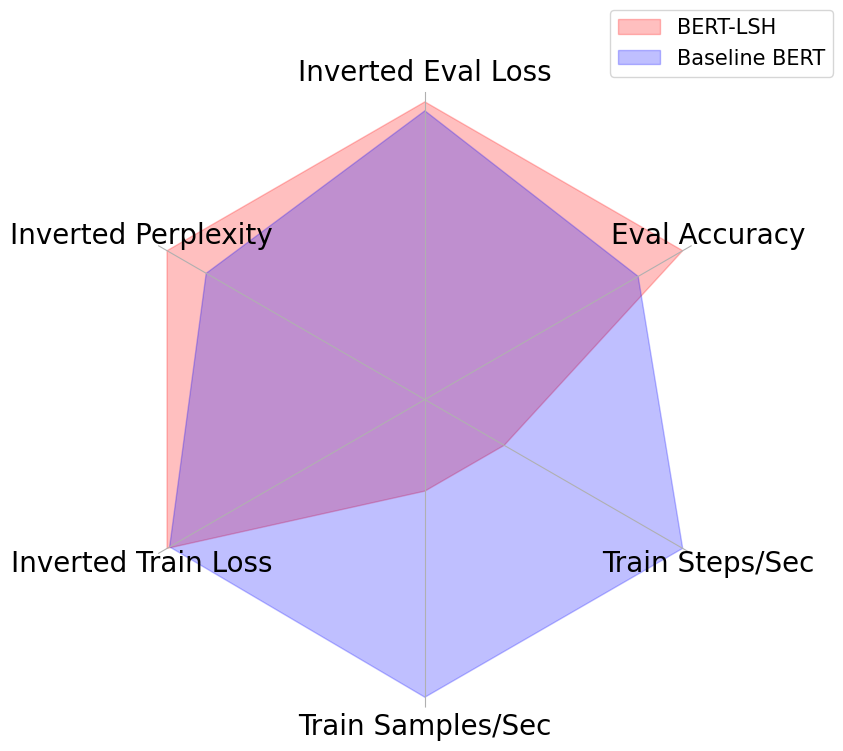}
    \caption{\textbf{Training Radar Plot}: A plot for the training parameters and metrics. Values further away from the center represent "better" performance.}
    \label{sst-finetune}
\end{figure}

\subsection{GLUE SST-2 Fine-tuning Results}
After the finetuning phase, the results indicated a nuanced performance difference between the \texttt{BERT-LSH} and the baseline \texttt{BERT} models. During the training iterations, the baseline \texttt{BERT} model exhibited a more rapid decrease in training loss and reached a lower final training loss compared to the \texttt{BERT-LSH} model \ref{sst-finetune}. This typical behavior would usually suggest a more effective learning process for the baseline BERT. However, the scenario took an intriguing turn when considering the performance on the test set. \texttt{BERT-LSH} achieved a slightly lower loss on the test set than the baseline BERT, which was an unexpected yet enlightening outcome.

\begin{figure}[htp]
    \centering
    \includegraphics[width=\linewidth]{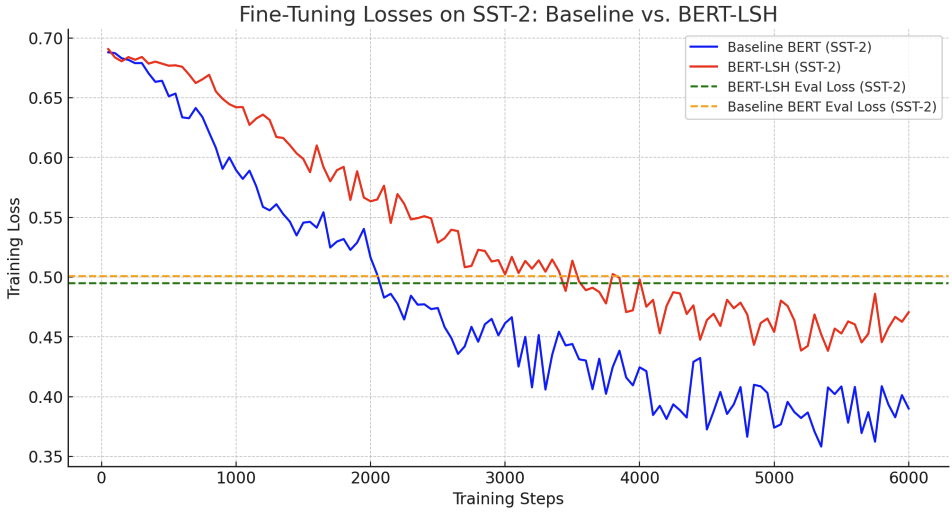}
    \caption{\textbf{Text Classification Finetune Loss}: \texttt{BERT-LSH} showed worse training loss but better evaluation loss than when compared to the Baseline \texttt{BERT} model during finetuning on SST-2 dataset}
\end{figure}
\FloatBarrier

This divergence between training loss and test loss is particularly noteworthy. Despite its higher training loss, the \texttt{BERT-LSH} model's superior performance on the test set implies that it has a stronger generalization capability. This is an indication that the \texttt{BERT-LSH} model is learning a more robust representation of the data. It seems to be capturing the essence of the underlying distribution rather than memorizing the training data, a common pitfall known as overfitting. However, \texttt{BERT-LSH}'s test set accuracy is 7\% lower than baseline \texttt{BERT}, which is an interesting result that prompts for more experimentation. One reason of why this happened could be that \texttt{BERT-LSH} is underfitting the training data, as indicated by the higher training loss.

\begin{table}[ht!]
\centering
\scriptsize 
\begin{tabularx}{\columnwidth}{Xcc} 
\hline
\textbf{Metric} & \textcolor{blue}{\texttt{Baseline BERT}} & \textcolor{red}{\texttt{BERT-LSH}} \\
\hline
Evaluation Loss                       & 0.5009        & \textbf{0.4948}            \\
Evaluation Accuracy &  \textbf{0.7672} & 0.7603 \\
Evaluation Runtime(s)                & \textbf{0.3263}        & 2.7963            \\
Evaluation Samples/s         & \textbf{2672.3}        & 311.846           \\
Evaluation Steps/s           & \textbf{42.904}        & 5.007             \\
Total Time(s)               & \textbf{125}           & 1110     \\
\hline
\end{tabularx}
\caption{Fine-Tuning Benchmarks for Baseline \texttt{BERT} and \texttt{BERT-LSH} Models on SST-2 }
\label{tab:fine_tuning_benchmarks}
\end{table}
\FloatBarrier

The observed resistance to overfitting suggests that the \texttt{BERT-LSH} model, with its LSH-based attention mechanism, is potentially focusing on the most salient features in the data which are more generalizable. This characteristic could be particularly beneficial in real-world scenarios where models often encounter data that differ from the examples seen during training. The ability of \texttt{BERT-LSH} to maintain a lower loss on the test set, despite a higher training loss, points to a model that is not only robust in the face of noise or variations in the data but also adept at handling the complexities of language beyond the confines of the training corpus. These findings merit further exploration into the model architecture and training dynamics to fully understand and harness the generalization benefits observed with BERT-LSH.

\subsection{SQuAD Fine-tuning Results}

Finetuning on the SQuAD dataset revealed some compelling insights. After three epochs, equivalent to 7500 training steps, the training loss plateaued after approximately 2500 steps, suggesting that the model had ceased to gain further knowledge from the data. This leveling off of the learning curve typically indicates that the model has reached its learning capacity under the current configuration.

\begin{figure}[htp]
    \centering
    \includegraphics[width=\linewidth]{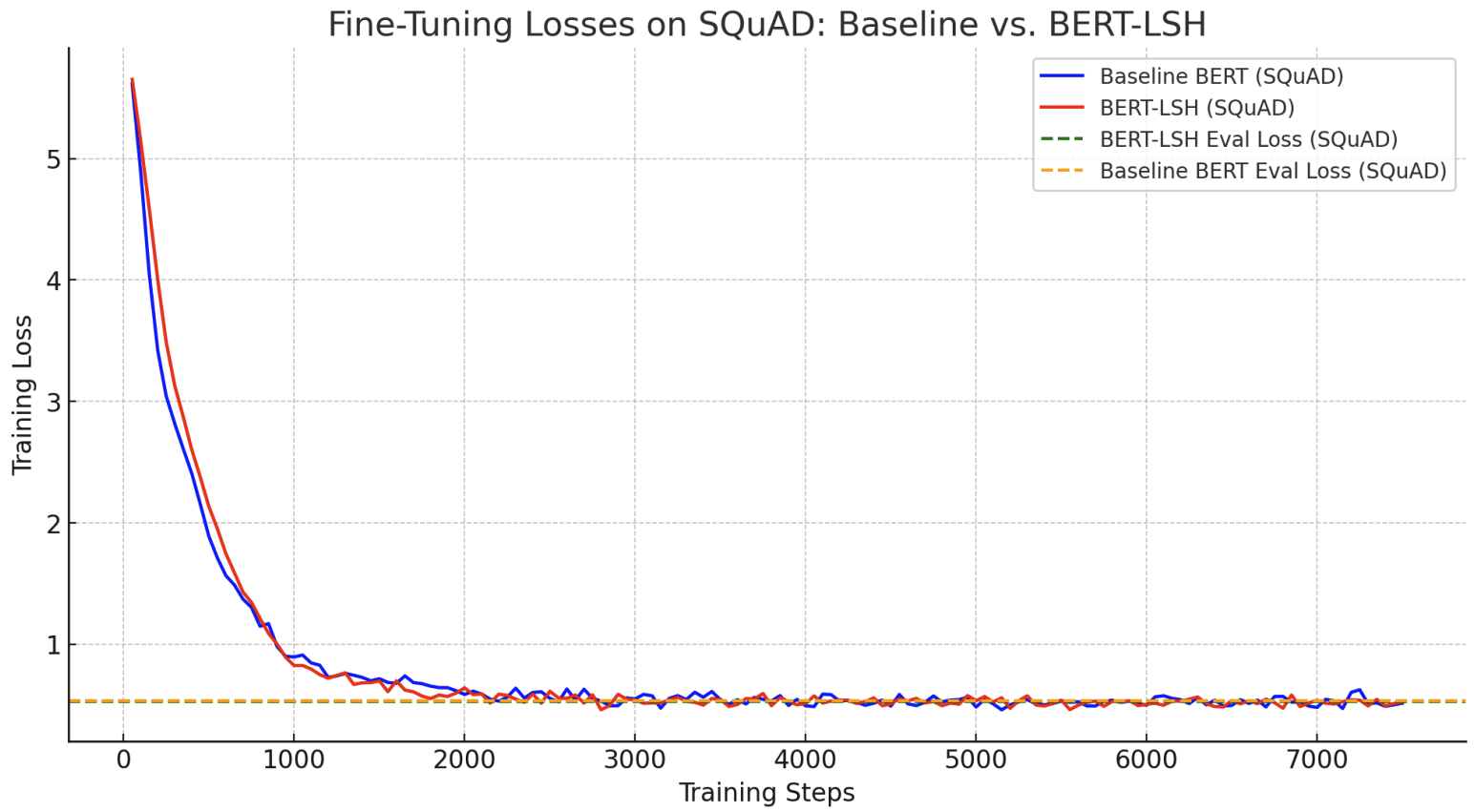}
    \caption{\textbf{Question Answering FineTune Loss}: SQuAD2.0 dataset}
\end{figure}

Despite this, both \texttt{BERT-LSH} and the baseline \texttt{BERT} models concluded their training with very similar losses on both the training and test sets. The final evaluation showed a marginal difference in performance, with \texttt{BERT-LSH} achieving a slightly lower evaluation loss of 0.53116 compared to the baseline BERT's 0.53637, hinting at a minor edge in test set generalization for BERT-LSH.

\begin{table}[h]
\centering
\scriptsize 
\begin{tabularx}{\columnwidth}{Xcc} 
\hline
\textbf{Metric} & \textcolor{red}{\texttt{Baseline BERT}} & \textcolor{blue}{\texttt{BERT-LSH}} \\
\hline
Evaluation Loss                    & 0.5364                 & \textbf{0.5312}            \\
Evaluation Runtime (s)             & \textbf{2.90}          & 28.80             \\
Evaluation Samples/s      & \textbf{3646.85}       & 366.97            \\
Evaluation Steps/s        & \textbf{57.27}         & 5.76              \\
Total Training Time (s)            & \textbf{120.91}        & 1140              \\
\hline
\end{tabularx}
\caption{Fine-Tuning Benchmarks on SQuAD2.0 for Baseline \texttt{BERT} and \texttt{BERT-LSH} Models}
\label{tab:fine_tuning_benchmarks_squad}
\end{table}
\FloatBarrier

\section{Conclusion}

Our exploration into the \texttt{BERT-LSH} model has yielded promising results, particularly in the realm of computational efficiency. By incorporating an LSH-based attention mechanism, our model has demonstrated a marked reduction in the absolute compute required, as measured in KFLOPs and the number of dot products, when compared to the traditional full self-attention mechanism. This reduction does not come at the cost of downstream tasks; \texttt{BERT-LSH} has shown the ability to maintain, and surprisingly surpass, the evaluation metrics of the baseline \texttt{BERT} model across various benchmarks in pretraining and finetuning.

Despite progress, \texttt{BERT-LSH}'s current form isn't ready for commercial or industrial deployment due to scalability and integration challenges. The practical applicability of \texttt{BERT-LSH} is contingent upon further refinement to ensure that it can operate within the complex and demanding environments that characterize real-world applications. 

Nonetheless, the efficiencies we have observed—particularly in terms of computational resources required for processing—underscore the potential of LSH to revolutionize the accessibility of transformer models. By reducing the compute overhead, LSH opens the door to deploying advanced NLP models in scenarios where such resources are scarce or where energy efficiency is paramount.

There are many further work that can be done on BERT-LSH. This will involve optimizing the LSH algorithm for better synergy with parallel computing architectures, which are increasingly commonplace in the industry. Moreover, a deep dive into the generalization abilities of the model will be critical to understand its performance nuances fully.

\nocite{*}
\printbibliography

\end{document}